\documentclass[journal]{IEEEtran}

\usepackage{setspace}
\usepackage{cite,graphicx,amsmath,amssymb}
\usepackage{subfigure}
\usepackage{citesort}
\usepackage{fancyhdr}
\usepackage{mdwmath}
\usepackage{mdwtab}
\usepackage{balance}
\usepackage{xcolor}
\usepackage{bm}
\usepackage{amsthm}
\usepackage{threeparttable}
\usepackage{algorithm}
\usepackage{algorithmic}
\usepackage{multirow}
\usepackage{flafter}

\begin{document}
\title{Robotic Communications for 5G and Beyond: Challenges and Research Opportunities}

\author{
Yuanwei~Liu, Xiao~Liu, Xinyu~Gao, Xidong~Mu, Xiangwei~Zhou, Octavia A.~Dobre, and H. Vincent~Poor

\thanks{This work has been submitted to the IEEE for possible publication. Copyright may be transferred without notice, after which this version may no longer be accessible.}
\thanks{Y. Liu, X. Liu, and X. Gao are with Queen Mary University of London; X. Mu is with Beijing University of Posts and Telecommunications; X. Zhou is with Louisiana State University; O. A. Dobre is with Memorial University; H. V. Poor is with Princeton University.}

}
\maketitle

\begin{abstract}
The ongoing surge in applications of robotics brings both opportunities and challenges for the fifth-generation (5G) and beyond (B5G) of communication networks. This article focuses on 5G/B5G-enabled terrestrial robotic communications with an emphasis on distinct characteristics of such communications. Firstly, signal and spatial modeling for robotic communications are presented. To elaborate further, both the benefits and challenges derived from robots' mobility are discussed. As a further advance, a novel simultaneous localization and radio mapping (SLARM) framework is proposed for integrating localization and communications into robotic networks. Furthermore, dynamic trajectory design and resource allocation for both indoor and outdoor robots are provided to verify the performance of robotic communications in the context of typical robotic application scenarios.
\end{abstract}


\section{Introduction}

Driven by advanced manufacturing processes, robots have evolved into versatile machines that have significant success to date~\cite{IEEEhowto:Siegwart}. In recent years, the increasing difficulty and complexity of robotic tasks have motivated an increased focus on robotic communication systems, which can play a decisive role in the performance of robotic systems. Particularly, when high dimensional data such as video characteristics is gathered and processed, the communication link established between a cloud server and robotic systems, can mitigate the cost of on-robot hardware. Additionally, emerging wireless networks with high data rates and stability can make the delay incurred by transmission negligible, which would allow robotic systems to complete increasingly complex missions. As a result, networked robotic systems are of significant interest. In view of these benefits, in this paper we focus on the application of the fifth-generation (5G) or beyond (B5G) networks to robotic systems. The features of high data speed, low latency, and reliable stability in 5G or B5G networks offer opportunities to enable breakthroughs in networked robotic systems. We begin our discussion by distinguishing different categories of terrestrial robots and their corresponding application scenarios.

\begin{figure*}
  \centering
  \includegraphics[width=4in]{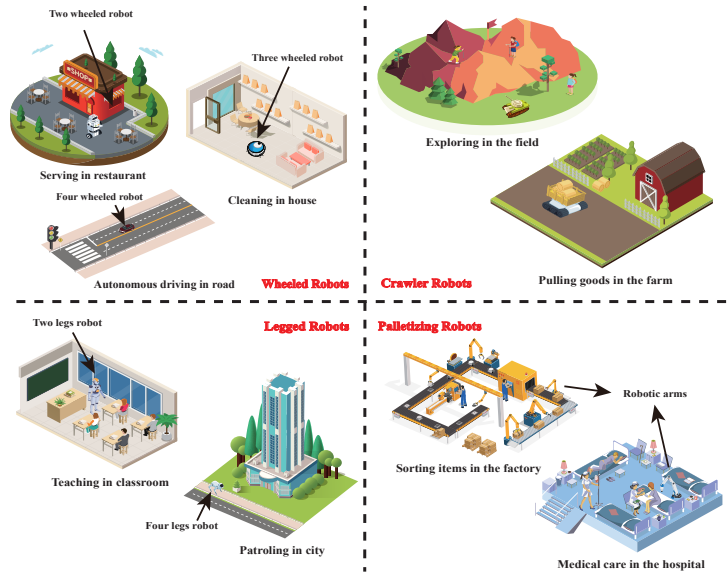}
  \caption{Different categories for terrestrial robots in specific application scenario.}
  \label{ground_robot_categories}
\end{figure*}

In light of distinct movement modes for terrestrial robots, as shown in Fig.~\ref{ground_robot_categories}, they can be broadly classified into four categories~\cite{IEEEhowto:Thrun}: wheeled robots, crawler robots, palletizing robots, and legged robots. Wheeled robots and crawler robots rely on wheels to move, the difference is reflected in the turning radius of the crawler robots that can be regarded as half of their own length. Additionally, wheeled robots are adopted on the hard ground (e.g., cement roads, asphalt roads, tile roads), while crawler robots are utilized in the soft road (e.g., mud roads, wetlands). Palletizing robots are a kind of robotic arms that are widely used in industry. Legged robots are quite complex, with sophisticated leg control systems. However, compared to the other three types, they can execute the greatest variety of tasks, including human-computer interactions in classroom, exhibition halls, etc.

Against the above background, we focus on 5G/B5G-enabled terrestrial robotic communications. The new contributions of this paper are that we highlight the signal and spatial modeling for terrestrial robotic communications, as well as the integration of localization and communications into robotic networks. Finally, we discuss the dynamic trajectory design and resource allocation for terrestrial robots by proposing the robotic-to-infrastructure (R2I) communication framework.

\section{Motivation and Applications of Integrating Robotic Communications into 5G and B5G}

Compared to conventional wireless networks, robotic communication networks have the following characteristics:

\begin{itemize}
  \item \textbf{Integrated Localization and Communication.} Robotic localization is essential for robots to complete certain tasks, which also provides a solid foundation for the intelligence of robots. Existing approaches, however, tend to stress localization at the expense of hardware complexity, while cellular assistance strategies that can simultaneously ensure efficient localization and reduce hardware costs have not been investigated in depth. Thus, integrating communication and localization is desirable, for which the limitations and challenges can be summarized as follows: 1) Working robots may enter communication dead zones. 2) The approach for the establishment of communications between base stations (BSs)/access points (APs) and robots arouse the localization accuracy for robots since errors accumulation and sky-high cost by built-in sensors in them. 3) The quality of communications between an BS/AP and robot determines how much communications can be traded for hardware complexity on the robot.
  \item \textbf{Dynamic Self-Decision Making.} In contrast to conventional communication networks, in 5G/B5G-enabled robotic communication networks, robots have to empower themselves with decision-making ability for rapidly adapting to their dynamic environments\cite{Shang2019Unmanned}; this indicates that robots have to make decisions without the intervention of humans by learning from the unknown environment to instantaneously adjust their control policies. Moreover, the decision policy has to aim for the farsighted network evolution instead of myopically striking the current benefits in the conventional communication networks.
  \item  \textbf{Heterogenous Mobility and Heterogeneous Quality-of-Service (QoS) Requirements.} In conventional communication networks, when considering the motion of mobile users, user mobility is usually ignored, and their time-variant QoS requirements are simplified as a constant value; this naturally reduces the dynamic scenario to a static one. However, when considering heterogenous mobility/QoS requirements in robotic communication networks, dynamic trajectory design and resource management introduce more complex parameters to be optimized simultaneously compared to static communication networks.
  \item  \textbf{Low-latency Requirements for Remote Control of Unmanned Robotics.} Since 5G/B5G-enabled robotics are danger-intolerant and delay-sensitive, uninterrupted and ubiquitous connectivity is necessary in order to provide ultra-reliable low-latency communications for terrestrial robots\cite{Yuan2018Ultra}. This is essential so that collisions can be avoided and general operating safety can be provided.
\end{itemize}

\begin{figure*}
\centering
\includegraphics[width=4.5in]{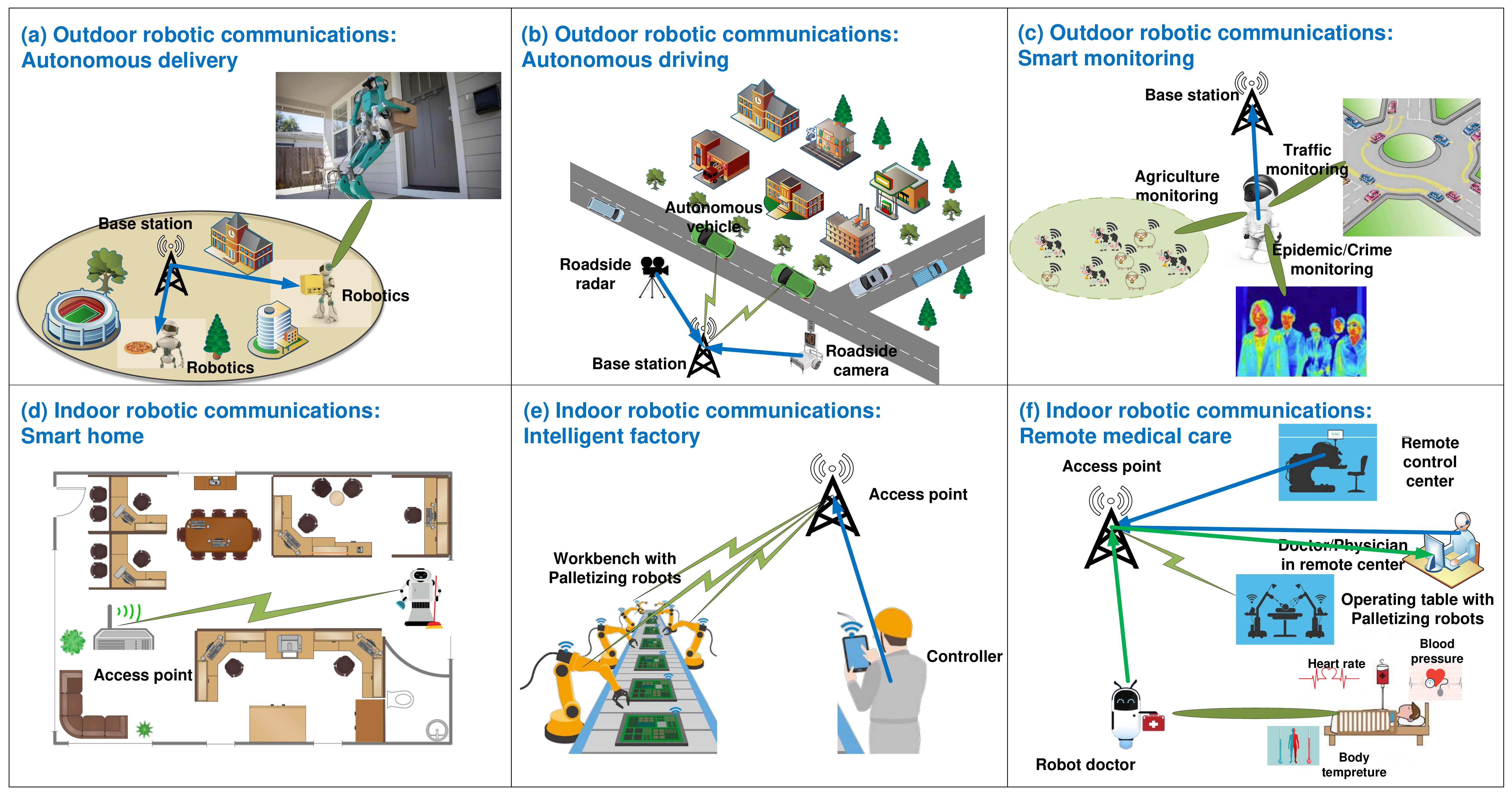}
\caption{Terrestrial robots and their applications for 5G and B5G.}\label{robotic}
\end{figure*}

As mentioned above, 5G/B5G-enabled robotic communication networks have been the focal point of the automotive research field for a while. Fig.~\ref{robotic} illustrates three outdoor robotic communication scenarios and three indoor robotic communication scenarios. More specifically, the first outdoor robotic communication network characterizes the application of 5G/B5G networks in autonomous delivery scenario. Industry companies such as Amazon, DHL, Jingdong, and UPS, have already employed autonomous robots to deliver commodities in a time and energy-saving manner\cite{Wen2018Swarm}. In the second outdoor robotic communication network, a cellular-connected autonomous driving scenario is discussed\cite{Yaqoob2019Autonomous}, where R2I communications based on roadside unites are considered instead of costly onboard sensors and radars. The third outdoor robotic communication network illustrates the smart monitoring scenario based on terrestrial robots. Robots can be employed at the intersection to monitor the traffic information and send the collected information to the control center. Further, robots can be deployed in an open public area to take pedestrians' body temperatures or other biological samples for monitoring pandemic, and they can also have the function of facial recognition for monitoring crime. Additionally, robots can act as shepherd dogs by enabling them with decision-making ability. The first indoor robotic communication network demonstrates 5G/B5G-enabled home robotic communications, such as floor mopping robots, robotic waiters in the restaurant, as well as robotic ushers in the shopping mall. These robots need to design their trajectories in the dynamic indoor environment for guaranteeing the communication link between them and BSs/APs. Additionally, physical collisions among terrestrial robots or between robots and obstacles have to be avoid no matter how the environment changes. In the second indoor robotic communication network, robots can replace human in missions that are dangerous in the intelligent factory. Additionally, with the aid of palletizing robots, manufacturing can be more efficient and without interruption. The third indoor robotic communication network illustrates a remote surgery, which can be operated safely and smoothly by palletizing robots with the aid of cellular networks. Thus, physicians in hospitals of highly urbanized areas can perform an operation on patients in areas where Internet cables are difficult or cannot be laid. Moreover, mobile robots in intelligent hospitals are capable of being robot doctors for remote medical caring and monitoring by collecting biological samples/data of patients, such as heart rate, body temperature, and blood pressure.

\section{Signal and Spatial Modeling for Robotic Communications}
In this section, we first introduce the signal modeling for incorporating robotic users in 5G/B5G wireless networks, and then discuss the distinct spatial modeling for evaluating the fundamental performance limits of robotic communications.

\subsection{Signal Modeling}

The channel model between the controllers (e.g., BSs or APs) and robots in general follows the well investigated conventional terrestrial transmission, and existing literature on channel modeling and field test can be referred~\cite{IEEEhowto:3GPP}. However, the robots also offer unique characteristics which can be exploited for facilitating the applications of robotic communications. More specifically, in conventional communication networks, the mobility of the human users is generally in a random and uncontrollable manner. By contrast, the mobility of robots is artificially controllable or can be predicted with high accuracy. Driven by this feature, it is naturally to design the trajectory of mobile robots and the deployment location of fixed robots with the aim of establishing a line-of-sight (LoS) link with the BSs/APs. As a result, it is suitable to use the Rician fading model with a specific Rician factor for robotic communications, which characterizes the deterministic LoS components and the random non-LoS (NLoS) components caused by signal scattering, reflection, and diffraction on surrounding objects\cite{Mu2020Intelligent}. For different application scenarios, the Rician factor should be carefully selected for striking a balance between the LoS and NLoS channel components. For example, for urban environments with high-rising buildings and crowded pedestrians, it is almost impossible for the robots to have a LoS link. Thus, a relatively small or even zero Rician factor is preferable in practice. On the other hand, for suburban or indoor big factory environments, the channel is characterised by a large Rician factor due to the vanishing signal blockage. In this case, the large-scale fading (e.g., the distance-dependent path loss and LoS component) is dominated and can be known when carrying out the robots' trajectory/deployment design, which reduces the overhead for channel acquisition.

Despite the above benefits, a major challenge of robotic communications is that, compared to human users, the Doppler effect of robotic channels can be more severe, especially for robots traveling at high speeds, such as autonomous driving vehicles. The situation could be even worse when the millimeter-wave bands is considered for robotic communications, which requires further investigation on how to deal with the resulting Doppler effect.
\subsection{Spatial Modeling}

The mobility of robots is known to have a fundamental impact on the performance of robotic networks. Stochastic geometry, recognized as a powerful tool for capturing the spatial randomness and the mobility of wireless networks, has experienced a significant amount of research in recent years. In contrast to conventional wireless communications, the mobility characteristic of robotic networks is an essential point to be considered. As mentioned above, compared to the mobile users in wireless networks whose mobility behaviors are unpredictable, the mobility behaviors of the robotic users are usually predefined. As a result, the mobility assumptions of stochastic geometry used in conventional networks are no longer accurate for robotic networks as the robots can either act as BSs/APs or users. Inspired by this characteristic, incorporating the robots' mobility in robotic networks is a promising research direction and the related research is still in its infancy. Another motivation of investigating the spatial modeling is that the robots are widely used in indoor environment in dense urban networks, while current research contributions of stochastic geometry mainly focus on the large-scale outdoor scenarios. In the design of indoor stochastic-geometry based robotic networks, several key factors need to be considered: 1) in contrast to outdoor scenarios which are typically two-dimensional  (2D), an accurate three-dimensional spatial model is required for indoor robotic communications; 2) the blockage model needs to be reconsidered for indoor robotic communications; and 3) the flexible user association schemes need to be reinvestigated as the received signals of indoor robotics are heavily dependent on the combinations of LoS and NLoS channels. Thus, the well-adopted nearest user association or average-received power-based user association schemes\cite{Andrews2012SG} in conventional networks are not efficient any more.

\begin{figure*}[htbp]
  \centering
  \includegraphics[height=2.8in,width=3.6in]{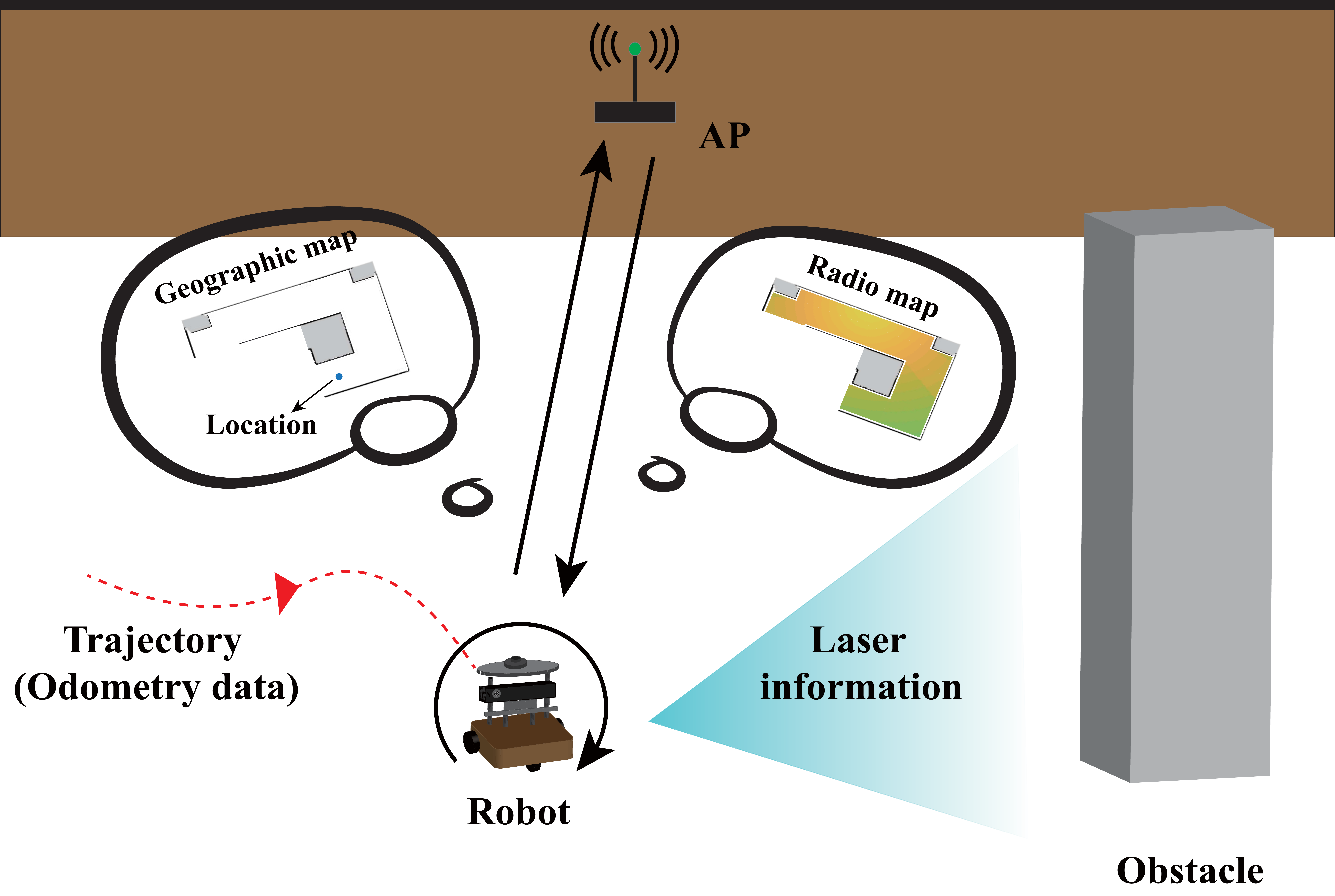}
  \caption{The process for SLARM framework.}
  \label{SLARM_core}
\end{figure*}

\section{Integrating Localization and Communications For Robots}

Localization is the course of events that robots percept and determine their locations in the environment, which is the premise for completing designated missions. Especially in environments where humans fail to probe beforehand, localization appears to be essential. For conventional mobile robots, the precise layout of the environment obtained by localization can guarantee the avoidance of physical collisions when accomplishing the assigned missions. However, the introduction of high-dimensional data, accumulation errors for data processing, and the sky-high cost of built-in sensors make localization non-trivial, which indicates that relying on themselves to overcome localization is far from enough for robots. Thus, the employment of communication-aided strategy provides possible solutions to handle these difficulties. There are two points of penetration for integrating localization and communications in robotic systems: 1) Communication-aided localization for robots, and 2) The combination of communication and robotic localization can evoke a basis for other missions (e.g., navigation). For the communication-aided localization, several communication nodes are pre-set in the environment and assist localization by the information transferred between them and robots. The main communication-assisted localization technologies are assisted by the global navigation satellite system (AGNSS), observed time difference of arrival (OTDOA) and enhanced cell ID (E-CID). AGNSS is a technology that combines BSs/APs information and satellite navigation information to locate mobile robots. Satellites above the robot capture information related to the robot's location, such as differential calibration information, and provide this auxiliary information to the robot through the cellular network. The robot can use this information as a priori knowledge to optimize the search, thereby improving the positioning performance. In OTDOA, the location of the robot is determined by detecting the time difference between the arrival of signals from multiple different BSs/APs. Therefore, this technology requires the environment to be equipped with at least three BSs/APs to complete localization. E-CID is a localization method based on the coverage of each cell, which adopts the known geographic information of the serving cell to estimate the location of the robot. The serving cell information can be obtained by updating the tracking area (TA), and integrate TA with angle of arrival (AoA) taking into account the timing advance and the direction of arrival on the basis of the location method, so as to achieve more accurate localization. For an unknown environment, combing the collected information of robots with pre-set communication nodes, which makes remedying the errors caused by uncertainty of nodes' locations possible and offers a solution for the localization. Therefore, we explore the integration of localization and communication technologies for robots that can serve for other missions and propose a novel simultaneous localization and radio mapping (SLARM) framework. The SLARM consists of geographic and radio map constructions~\cite{Gao2020SLARM}. The geographic map construction aims to explore the locations of all objects in the unknown environment, such as BSs/APs, robots, obstacles, i.e., preparing for the establishment of a communication link, while the radio map characterizes the knowledge of location-dependent radio propagations (e.g., signal-to-noise ratio (SNR), achievable communication rate, etc.) between BSs/APs and robots. Relying on these obtained information, other latency and/or reliability sensitive robotic applications can be smoothly facilitated, e.g., allowing robots to accomplish tasks in areas having desired signal conditions without colliding with obstacles. Different from conventional radio environment mapping (REM)~\cite{REM}, where the radio map is obtained by deploying multiple sensors to collect measurement data from \emph{a priori known} geographic environment, the proposed SLARM scheme can simultaneously explore the geographic map and the radio map in \emph{an unknown environment} employing only one robot. In the following, we introduce the details of constructing the geographic and radio maps in the SLARM.

\subsection{Localization in Geographic Map}

A geographic map constructed by robots can be considered as a series of grid-like sub-maps, where the resolution is adopted to represent the size of each sub-map. A higher resolution provides more accuracy for mirroring the information on the geographic map; the essential information can be lost when the resolution is lower. The executor of geographic map construction is generally identified as a robot, whose location is a key factor in determining the specific information of each sub-map. Thus, the accuracy of robotic localization is used for evaluating the geographic map construction, which also demonstrates that localization and geographic map building are complementary. Among others, simultaneous localization and mapping (SLAM) algorithms are considered to be efficient tools for striking a balance between localization and geographic map construction. Here, a laser-based SLAM algorithm is considered, whose idea is to estimate a function to determine the probability of each object appearing in certain locations in the environment, namely, the information probability distribution in the environment. Additionally, repeated sampling methods are invoked in the process of the robotic data collection, which uses the mean square error as the evaluation index for the SLAM accuracy. The layout of all obstacles can be estimated by the expected target probability distribution function obtained to complete the construction of the geographic map.

\subsection{Integrate Localization and Communications in SLARM}

For communication-guided robotic missions, localization by robots themselves is not enough, and communication quality is another essential factor. For example, working robots may get stuck in communication dead zones (e.g., regions with extremely poor channel conditions) and out of contact, even though no physical collisions happen. These unique limitations give rise to the difficulty of completing the missions. Thus, it is imperative for robots to collect radio information. However, as described in the communication characteristics, the radio propagations between BSs/APs and robots are not only determined by their locations but also the surrounding objects. Here, we introduce the concept of radio map which characterizes the location-dependent radio information. The linchpin for SLARM is to achieve \emph{simultaneous}, which can minimize the difficulty caused by the separation of localization and radio map establishment mentioned above. Specifically, when robots start to explore from a given initial location to construct the geographic map via localization, they also \emph{simultaneously} calculate and record the radio status of the current location according to the returned location information obtained in the localization algorithm.


\begin{figure}[htbp]
    \centering
    \includegraphics[height=2.5in,width=3.4in]{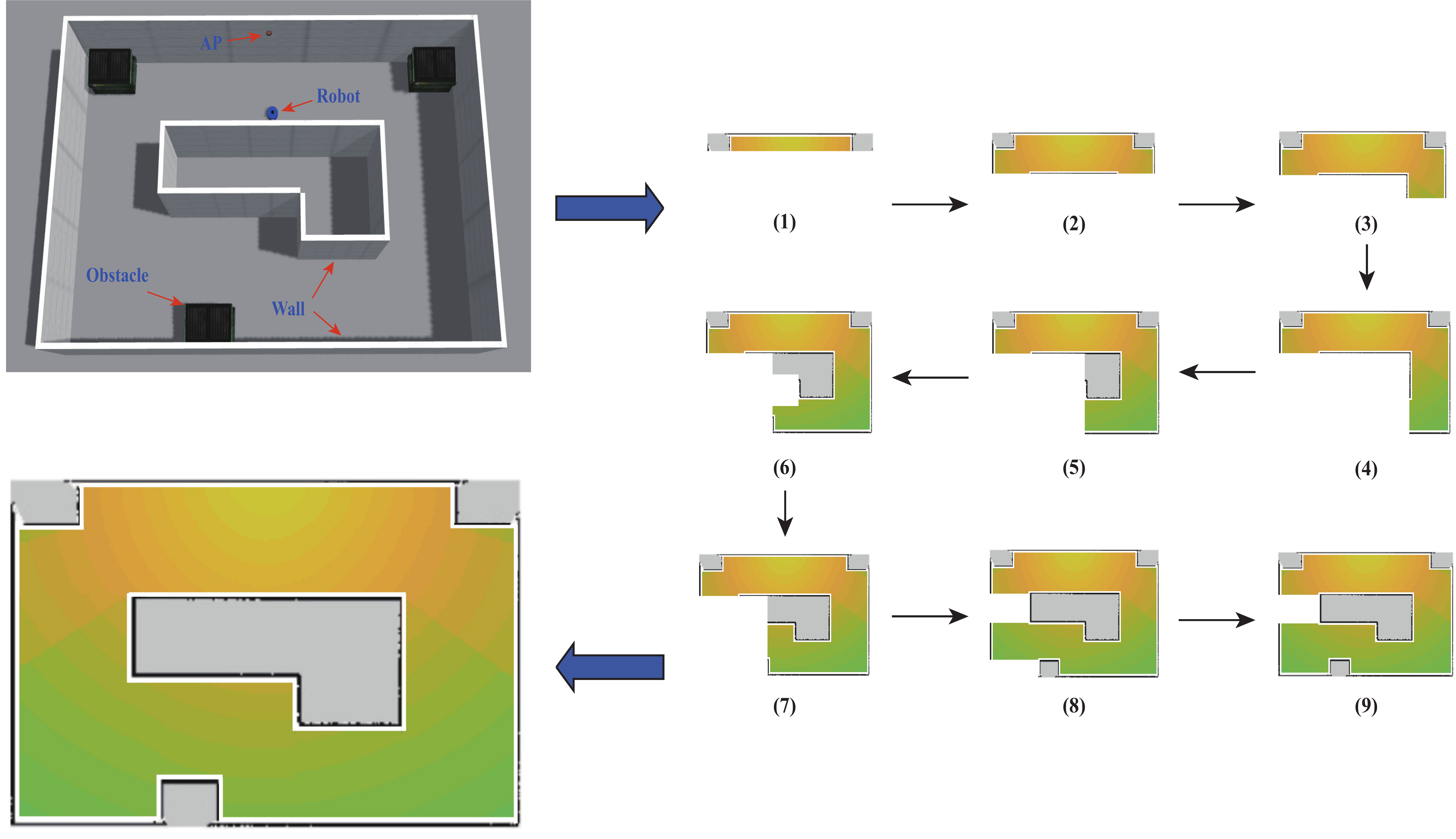}
    \caption{SLARM: geographic map and radio map are established simultaneously.}
    \label{SLARM}
\end{figure}

As shown in Fig.~\ref{SLARM_core}, the SLARM framework is partitioned into two parts. One is for localization, where odometry data and laser information are utilized as input parameters to obtain the results for the robotic localization and geographic map constructions. The other part is for the construction of the radio map, where the robotic location and location-dependent channel model are conceived as input parameters, while the required communication performance is used as an output result. In Fig.~\ref{SLARM}, a physical simulation environment with two-layer walls and three obstacles is considered, in which the height of each obstacle is much larger than that of robots. The black line in this figure represents the boundary between the obstacle and environment, and the gray area in this figure represents the impassable area for robots. For communication information, the closer the color of the passable area is to yellow, the better the channel gain is. On the contrary, the closer is to green, the smaller the channel gain is. In order to refine the SLARM framework, the glitches and irregularities due to sampling can be smoothed on a geographic map. We can also observe that the communication qualities (or channel gains) vary abruptly due to signal blockage. Thus, the obtained radio map can prevent robots from out of contact by indicating the distribution of channel conditions over the region of interest. Additionally, the space complexity of the proposed SLARM framework is mainly determined by the data size of the map, which grows with the square of the total number of grids on the map. Additionally, as one type of robots, unmanned aerial vehicles (UAVs) have been also widely considered for B5G communication network due to their high mobility and low cost \cite{Wu2020Cellular}. In particular, some studies have investigated trajectory design and resource allocation by integrating localization and communications into UAV-aided wireless networks The main requirements for our proposed framework are workable BSs/APs and movable robots, based on which the geographic map and radio map can be constructed simultaneously. Thus, the proposed framework can be adopted in UAV-aided wireless networks. The SLARM framework demonstrated in this paper is merely utilized beneath the pre-defined plane height of the 2D workspace, which is proper for regular obstacles and robots. Otherwise, the vision technology needs to be introduced for recognizing the minimum distance between the whole body of robot and obstacles.
\section{Dynamic Trajectory Design and Resource Allocation}

In contrast to conventional wireless networks, 5G/B5G-enabled robotic communication networks are characterized by more rapidly fluctuating network topologies and more vulnerable communication links. Hence, robots operate in a highly dynamic environment, which makes it challenge for trajectory design for robots to attain long-term benefits. Moreover, the specific time-varying data demand of each robot may be readily accommodated by robotic networks. Thus, the amount of required wireless resources (e.g., bandwidth, transmit power, and computational resource) also varies, which emphasizes the importance of agile resource allocation for robots.

\subsection{Dynamic Trajectory Design and Resource Allocation for Outdoor Robots}

In recent research contributions, autonomous vehicles are regarded as terrestrial robots. In the same manner as with vehicle-to-infrastructure (V2I) communications, R2I communications support the interaction amongst outdoor robots, shared roadside units (RSUs), roadside base stations (RBSs), pedestrians, vehicles and other local paraphernalia to enhance the operational safety and efficiency of outdoor robots. In R2I-assisted autonomous driving, the RSUs are used to either complement or replace the costly onboard units (OBUs) for collecting real-time traffic information, which facilitates the interaction between robots and their environments. The collected traffic information is sent from the RSUs to RBSs via the Internet. The RBSs also adopt a high-performance processor for processing both information and data. Thus, outdoor robots are capable of receiving reliable real-time traffic information, as well as noticing the presence of obstacles from the RBSs in NLoS scenarios where robots cannot directly sense.

\begin{figure}
\centering
\includegraphics[width=3in]{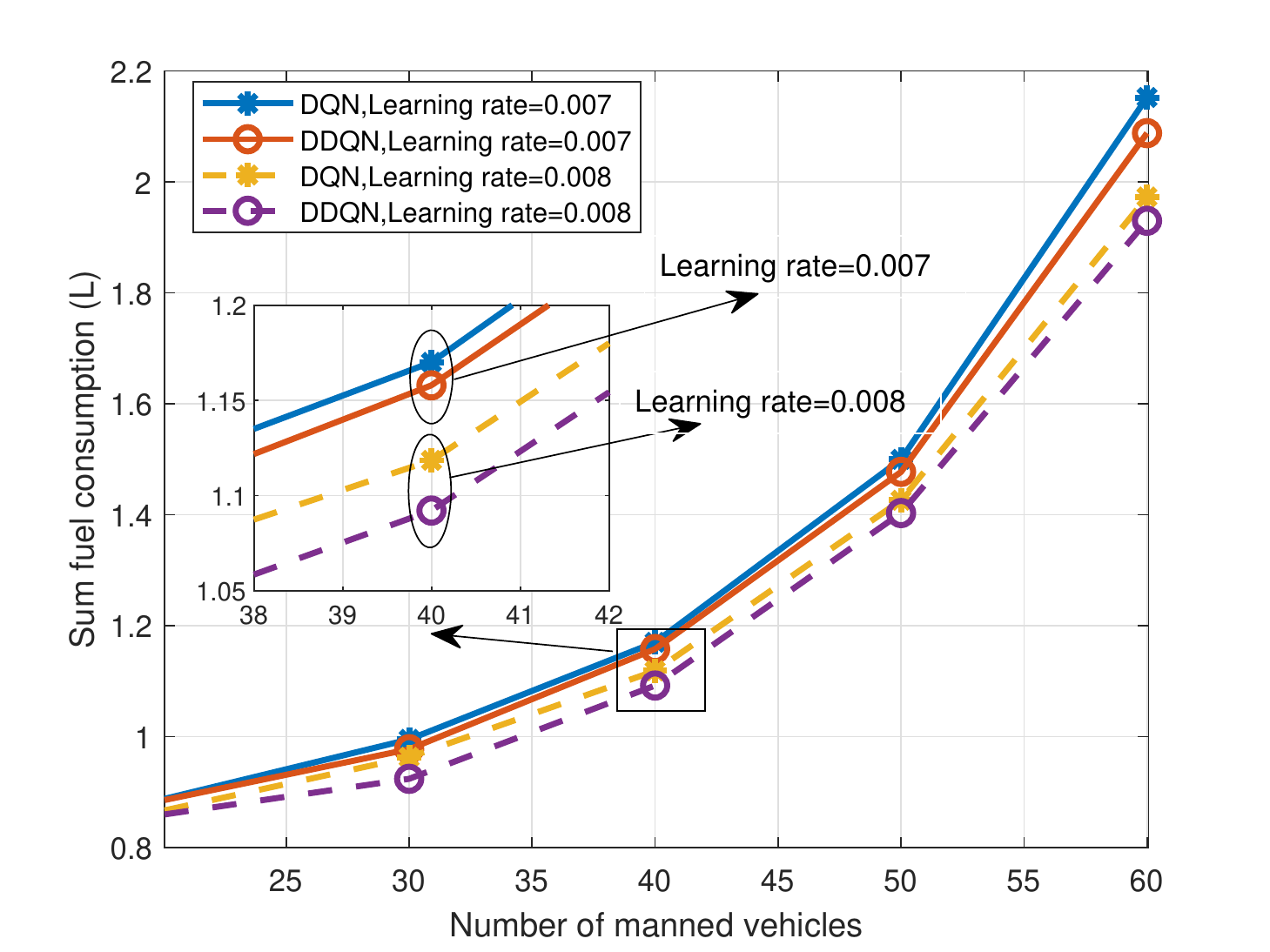}
\caption{Total fuel consumption vs the traffic conditions\cite{Xiao2020Enhancing}.}\label{fuelvsdensity}
\end{figure}

The conventional solutions are not capable of realizing the interaction between robots and their environments, as well as the cooperation between different robots, which indicates that robots cannot learn from the unknown environment for instantaneously adjusting their design policies. Additionally, in conventional solutions, the control policies of robots aim for myopically striking the current benefits for the networks without considering the farsighted network evolution. In an effort to overcome the aforementioned limitations when adopting conventional approaches in 5G/B5G-enabled robotic communication networks, machine learning (ML) approaches can be invoked to tackle the dynamic/uncertain environments of robotic in robotic networks. We focus our attention on reinforcement learning (RL) approaches, as they are capable of enabling robotics to learn from the real-time feedback of dynamic/uncertain environments, as well as from their historical experiences\cite{Challita2020ML}. Thus, RL approaches have been regarded as potential candidates for solving challenges encountered in 5G/B5G-enabled robotic communication networks. In RL-empowered robotic networks, robots are capable of rapidly adapting their control policies to their dynamic environments and continuously improving their decision-making ability.

\begin{figure}
\centering
\includegraphics[width=3in]{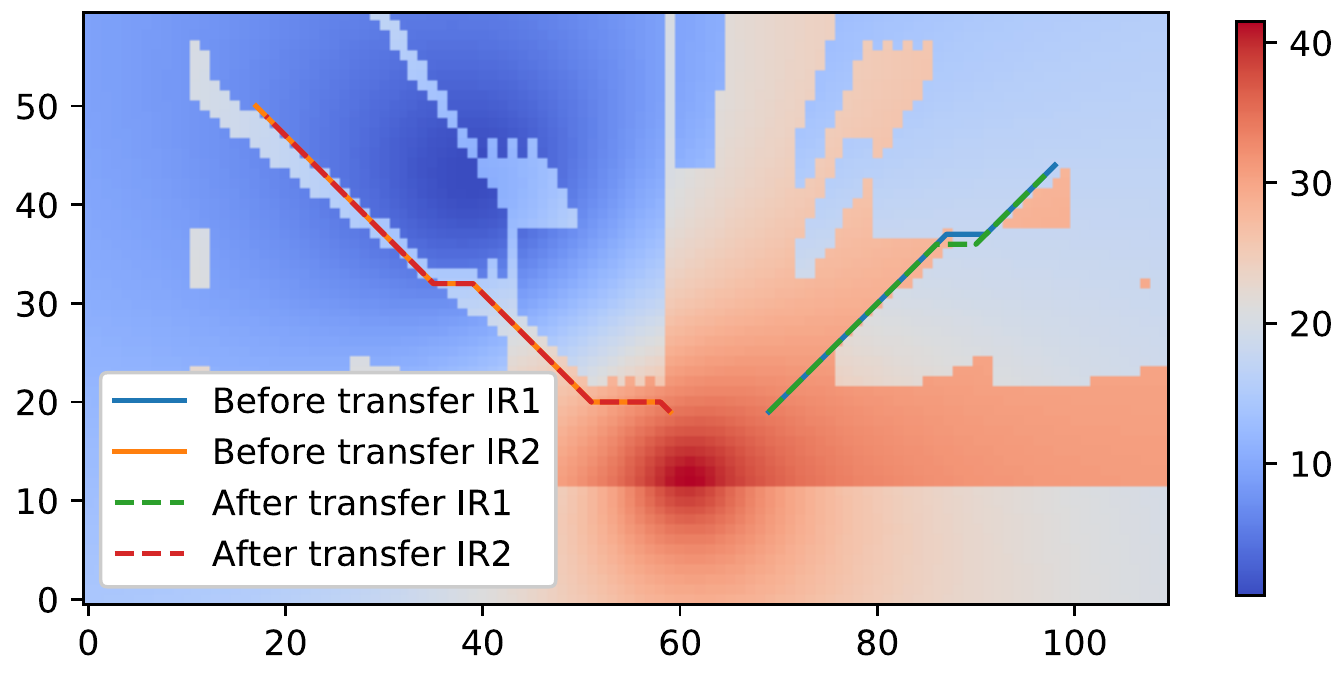}
\caption{Designed trajectories of robots on the radio map\cite{Zhong2020Path}.}\label{path2}
\end{figure}

Fig.~\ref{fuelvsdensity} illustrates the fuel consumption of the robot versus the density of manned vehicles. This figure indicates that by receiving real-time traffic information from BSs, the robot (acting as an agent in the proposed ML model) is capable of finding a policy for attaining both fuel economy and safe driving/operating. It can be observed that the fuel consumption of the robot increases rapidly as the density of manned vehicles raises. This is because as the number of manned vehicles increases, the robot needs to choose extra more actions (lane-change, acceleration and deceleration) for avoiding the collision.

\subsection{Dynamic Trajectory Design and Resource Allocation for Indoor Robots}

As mentioned above, robots are found to be extremely useful to replace humans or manned vehicles in indoor missions. In an effort to improve the efficiency of robots, they need to choose a time-saving path to the designated destination and maintain a qualified wireless link with the core network during the moving process in order to obtain the required information to display in real-time. Radio maps, which contain the propagation knowledge of the complex indoor environment, can be invoked for training the ML-aided dynamic trajectory and resource allocation model. Radio maps present the power distribution of the transmitted signal in a given area. A digital radio map can train the ML-aided dynamic trajectory and resource allocation model without the hardware consumption and physical venue expenditure.

Fig.~\ref{path2} characterizes the designed trajectories of robots. In an effort to avoid getting sequacious in the negative rewards and falling into a local optimum solution, the transfer learning (TL) is adopted. Both trajectories before and after the TL are compared in this figure. It can be observed that communication-connected robotics are capable of finding the optimal trajectory to the destination while guaranteeing the communication quality and avoiding collision. The trajectory planning before TL does not consider power allocation at each corresponding location since it is too complex for the agent to optimize the compound problem. Therefore, although the trajectories before the TL can reach the destination, the communication quality cannot be guaranteed.

In future ML-empowered robotic networks, the discrete trajectory design problem is coupled with the continuous resource allocation problem. How to design an RL model with both continuous and discrete state space is a potential research opportunity. Additionally, the design goals will be more ambitious than simply optimizing a single objective. How to design an RL model to obtain the Pareto-optimal solution of the multi-objective problem is another research opportunity.

\section{Conclusion}

In this article, 5G/B5G-enabled robotics communication networks have been considered. It has been argued that designing robotics communication networks is challenging due to heterogenous mobility and QoS, low-latency, and dynamic self-decision making requirements. The distinction of 5G/B5G-enabled robotics communication networks compared to conventional wireless networks has also been highlighted in terms of signal and spatial models. Additionally, integration between localization and communications in robotic networks has been considered by proposing a novel SLARM framework based on conventional localization and geographic map construction methods. Finally, case studies for both indoor and outdoor robots have been presented to verify the performance of 5G and beyond enabled robotics networks.

\bibliographystyle{IEEEtran}

 \end{document}